\documentclass[letterpaper, 10 pt, conference]{ieeeconf}  

\IEEEoverridecommandlockouts                              

\usepackage{graphicx}
\usepackage{cite}
\usepackage{amsmath,amssymb}
\usepackage{color}
\usepackage{booktabs}
\usepackage{url}
\usepackage{float}
\usepackage{algorithm}
\usepackage{algpseudocode}

\title{\LARGE \bf
BioLite U-Net: Edge-Deployable Semantic Segmentation for In Situ Bioprinting Monitoring
}

\author{Usman Haider$^{1,\dagger}$, Lukasz Szemet$^{1,\dagger}$, Daniel Kelly$^{2,3}$, 
Vasileios Sergis$^{2,3}$, Andrew C. Daly$^{2,3,\ddagger}$, and Karl Mason$^{1,\ddagger,*}$
\thanks{This work was supported by the University of Galway and funded by Research Ireland}
\thanks{$^{1}$School of Computer Science, College of Science and Engineering, University of Galway, Galway, H91 TK33, Ireland.}%
\thanks{$^{2}$CÚRAM – SFI Research Centre for Medical Devices, University of Galway, Galway, H91 W2TY, Ireland.}%
\thanks{$^{3}$Biomedical Engineering, School of Engineering, College of Science and Engineering, University of Galway, Galway, H91 HX31, Ireland.}%
\thanks{$^{*}$ Corresponding Author: karl.mason@universityofgalway.ie}%
\thanks{$^{\dagger}$ and $^{\ddagger}$These authors contributed equally.}%
}

\begin{document}

\maketitle
\thispagestyle{empty}
\pagestyle{empty}

\begin{abstract}

Bioprinting is a rapidly advancing field that offers a transformative approach to fabricating tissue and organ models through the precise deposition of cell-laden bioinks. Ensuring the fidelity and consistency of printed structures in real-time remains a core challenge, particularly under constraints imposed by limited imaging data and resource-constrained embedded hardware. Semantic segmentation of the extrusion process, differentiating between nozzle, extruded bioink, and surrounding background, enables in situ monitoring critical to maintaining print quality and biological viability.
In this work, we introduce a lightweight semantic segmentation framework tailored for real-time bioprinting applications. We present a novel, manually annotated dataset comprising 787 RGB images captured during the bioprinting process, labeled across three classes: nozzle, bioink, and background. To achieve fast and efficient inference suitable for integration with bioprinting systems, we propose a BioLite U-Net architecture that leverages depthwise separable convolutions to drastically reduce computational load without compromising accuracy. Our model is benchmarked against MobileNetV2 and MobileNetV3-based segmentation baselines using mean Intersection over Union (mIoU), Dice score, and pixel accuracy. All models were evaluated on a Raspberry Pi 4B to assess real-world feasibility. The proposed BioLite U-Net achieves an mIoU of 92.85\% and a Dice score of 96.17\%, while being over 1300× smaller than MobileNetV2-DeepLabV3+. On-device inference takes 335 ms per frame, demonstrating near real-time capability. Compared to MobileNet baselines, BioLite U-Net offers a superior trade-off between segmentation accuracy, efficiency, and deployability, making it highly suitable for intelligent, closed-loop bioprinting systems.

\end{abstract}

\section{Introduction}

Bioprinting is a transformative technology at the intersection of tissue engineering, regenerative medicine, and additive manufacturing \cite{murphy20143d, groll2016biofabrication, ozbolat2015bioprinting}. By precisely depositing cell-laden bioinks in a layer-by-layer manner, bioprinting enables the fabrication of complex three-dimensional (3D) biological structures, such as vascular networks, organoids, and scaffold-free tissues. As the field advances, maintaining high-resolution spatial fidelity and structural consistency becomes critical, not only for functional outcomes, but also for biological viability \cite{han2021study}.

Real-time monitoring of bioprinting offers significant promise in improving print quality, reducing failure rates, and enabling adaptive control \cite{he2016research, gudapati2016comprehensive, armstrong20201d, sergis2025situ}. However, achieving effective in situ monitoring faces two major challenges. First, the dynamic nature of extrusion, including nozzle movement, variable flow rates, and deposition behaviors, requires high spatial and temporal resolution to capture relevant features. Second, most bioprinters operate in constrained environments with limited computational resources, making it impractical to deploy conventional deep learning models that rely on high-end GPUs.

Semantic segmentation, which classifies each pixel of an image into semantic categories, has emerged as a powerful tool for fine-grained visual analysis. In bioprinting, segmenting the nozzle, bioink, and background can provide critical feedback to assess extrusion quality, detect anomalies, and enable closed-loop control \cite{kelly2025autonomous}. While segmentation has been widely adopted in fields such as autonomous driving and medical imaging, its use in bioprinting remains underexplored, particularly on resource-constrained embedded devices where minimizing inference times and model size are critical.

To address these challenges, we propose a lightweight semantic segmentation framework optimized for edge deployment in bioprinting environments. Our key contributions are as follows:

\begin{itemize}
    \item We introduce a new manually annotated dataset of 787 RGB images captured during bioprinting, labeled into three relevant classes: \textit{nozzle}, \textit{bioink}, and \textit{background}. This dataset fills a key gap in the literature by enabling semantic understanding of the bioprinting process.
    
    \item We design a novel \textbf{BioLite U-Net} architecture that uses depthwise separable convolutions to reduce computational complexity without compromising segmentation accuracy for real-time inference on low-power devices.
    
    \item We compare BioLite U-Net against MobileNetV2 and MobileNetV3-based segmentation backbones in terms of mIoU, Dice score, pixel accuracy, and inference latency on a Raspberry Pi 4, demonstrating a favorable trade-off between performance and efficiency.
\end{itemize}

To our knowledge, this is the first study to perform real-time semantic segmentation of bioprinting imagery on embedded hardware. Our work lays the foundation for smart, autonomous bioprinters capable of adaptive control and real-time feedback, ultimately contributing to the broader vision of scalable and precise tissue fabrication.

\section{Related Work}

Semantic segmentation has become increasingly important in the monitoring and control of additive manufacturing (AM) processes. Encoder-decoder architectures such as U-Net~\cite{ronneberger2015u} have been widely adopted for pixel-level understanding across many domains, from biomedical imaging to industrial inspection. In AM, segmentation models have been used to detect porosity, spatter, melt pool irregularities, and other layer-wise anomalies~\cite{deshpande2024review, xu2020review}. DeepLabv3+~\cite{chen2018encoder}, which introduced atrous spatial pyramid pooling to capture multi-scale context, has shown strong performance in high-precision segmentation tasks.

Despite these advances, segmentation has been underutilized in the bioprinting domain. Bioprinting presents unique challenges, such as the optical translucency of bioinks, the deformability of bioinks, and delicate biological constraints that limit imaging quality and consistency. Gugliandolo et al.~\cite{gugliandolo2024new} used infrared imaging to extract bioink filament profiles, but their approach was limited to thermal intensity thresholding. Ashley A Armstrong et al. \cite{armstrong2019direct} used a 2D laser displacement scanner to measure the deposited strands of bioink. These approaches provide classification or measurement heuristics but lack dense semantic labeling. Our work addresses this gap by proposing an end-to-end segmentation solution specifically designed for real-time bioprinting image streams. This builds upon prior work in the field, including classical computer vision for in-situ quality monitoring \cite{sergis2025situ} and CNN-based autonomous extrusion control \cite{kelly2025autonomous}, extending these efforts toward lightweight embedded deployment.

In recent years, lightweight segmentation models have been developed to meet the demands of embedded and resource-constrained systems. ENet~\cite{paszke2016enet} was an early real-time model that reduced computation by introducing early downsampling and bottlenecks. BiSeNet~\cite{yu2018bisenet} utilized a dual-path design to preserve spatial detail while maintaining high speed. MobileNet~\cite{howard2017mobilenets} and MobileNetV3~\cite{howard2019searching} introduced depthwise separable convolutions and neural architecture search for efficient image classification, and have since been widely adopted as backbones for segmentation models.

Efforts to further optimize U-Net have produced various lightweight U-Net architectures. Chen et al.~\cite{chen2024tinyunet} introduced a cascaded multi-receptive-field version of U-Net with significantly fewer parameters while maintaining high Dice scores in medical segmentation. Okman \emph{et al.}~\cite{okman2022l3unet} proposed L\textsuperscript{3}U-net, which runs real-time segmentation ($\approx$10 FPS) on low-power CNN accelerators like the MAX78000. JetSeg~\cite{lopez2023jetseg}, designed for Jetson Xavier, demonstrates strong real-time throughput with lightweight encoder-decoder blocks and optimized depthwise-dilated convolutions. Additionally, hardware-aware segmentation research has demonstrated promising performance on constrained platforms. Posso \emph{et al.}~\cite{posso2025uav_UNet} benchmarked a compact U-Net on CPU, GPU, and FPGA platforms, showing substantial latency improvements through hardware-specific optimization. Similarly, Kwon \emph{et al.} recently introduced HARD-Edge, achieving real-time (33 FPS) semantic segmentation on ARM Cortex-M microcontrollers~\cite{kwon2024hard}.

In summary, while classification and thresholding methods have been applied in bioprinting monitoring, real-time dense semantic segmentation remains largely unexplored. Our work builds on advances in lightweight CNNs and hardware-aware design principles to deliver a novel solution optimized for real-time segmentation in bioprinting environments.

\section{Methodology}

The objective of this work is to develop a lightweight and accurate semantic segmentation model capable of real-time performance on embedded systems during bioprinting. Our proposed framework consists of five main stages: data acquisition and preprocessing, segmentation using BioLite U-Net, postprocessing, and embedded deployment. 
\begin{figure*}[htbp]
    \centering
    \includegraphics[width=0.95\linewidth]{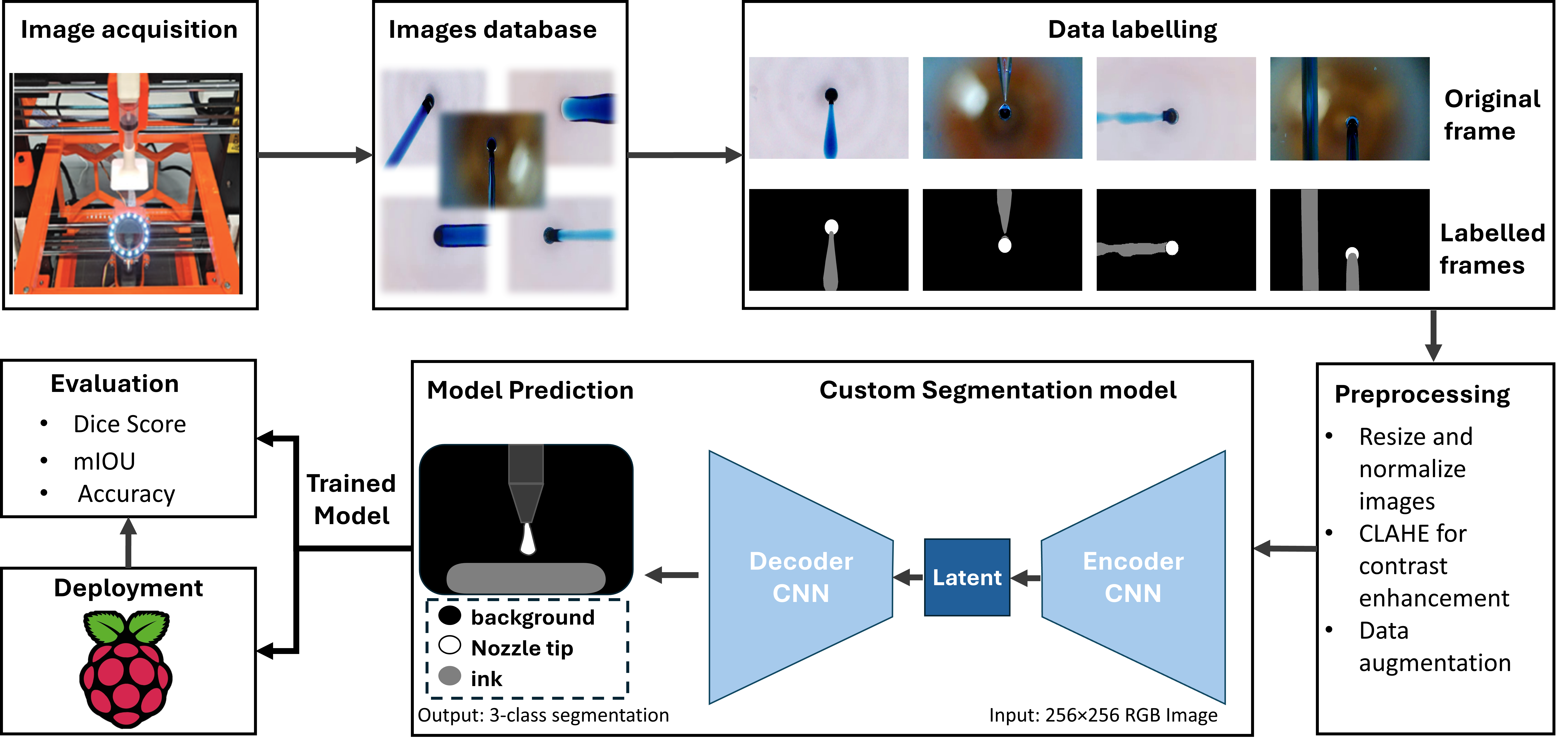}
    \caption{Overview of the proposed segmentation framework for real-time bioprinting monitoring.}
    \label{fig:framework}
\end{figure*}

As illustrated in Figure~\ref{fig:framework}, the system begins with a continuous video feed captured from a low-cost camera mounted on a bioprinter. Individual frames are extracted in real time and undergo a series of preprocessing steps, including:

\begin{itemize}
\item CLAHE-based \cite{reza2004realization} contrast enhancement to improve local visibility of ink traces.
\item Spatial resizing to a fixed resolution of $256 \times 256$ for network input.
\item Geometric and photometric augmentations (e.g., flipping, rotation, brightness jitter) to improve generalization during training.
\end{itemize}



\subsection{Segmentation Model}

Following preprocessing, the enhanced image frames are passed to the core segmentation engine, a custom BioLite U-Net, for pixel-wise classification into three semantic classes: background, ink trace, and nozzle. This segmentation step is crucial for downstream control decisions in bioprinting and thus requires both accuracy and speed sufficient for real-time deployment.

\subsubsection{BioLite U-Net Architecture}

At the core of our framework is a custom BioLite U-Net architecture, inspired by the original U-Net but significantly modified for efficiency. The architecture retains an encoder–decoder topology with skip connections~\cite{ronneberger2015u}, and introduces the following key optimizations:

\begin{itemize}
    \item \textbf{Depthwise Separable Convolutions}: All standard convolutional layers are replaced with depthwise separable convolutions~\cite{howard2017mobilenets}, which consist of a 3×3 depthwise convolution followed by a 1×1 pointwise convolution. This significantly reduces parameter count and computational load.
    
    \item \textbf{Shallow Encoding and Decoding}: The network contains few downsampling layers and few corresponding upsampling layers using bilinear interpolation, balancing feature richness and latency.
    
    \item \textbf{Skip Connections}: Feature maps from the encoder are concatenated with decoder feature maps at each resolution level to preserve spatial detail.
\end{itemize}

\begin{figure*}[htbp]
    \centering
    \includegraphics[width=0.9\linewidth]{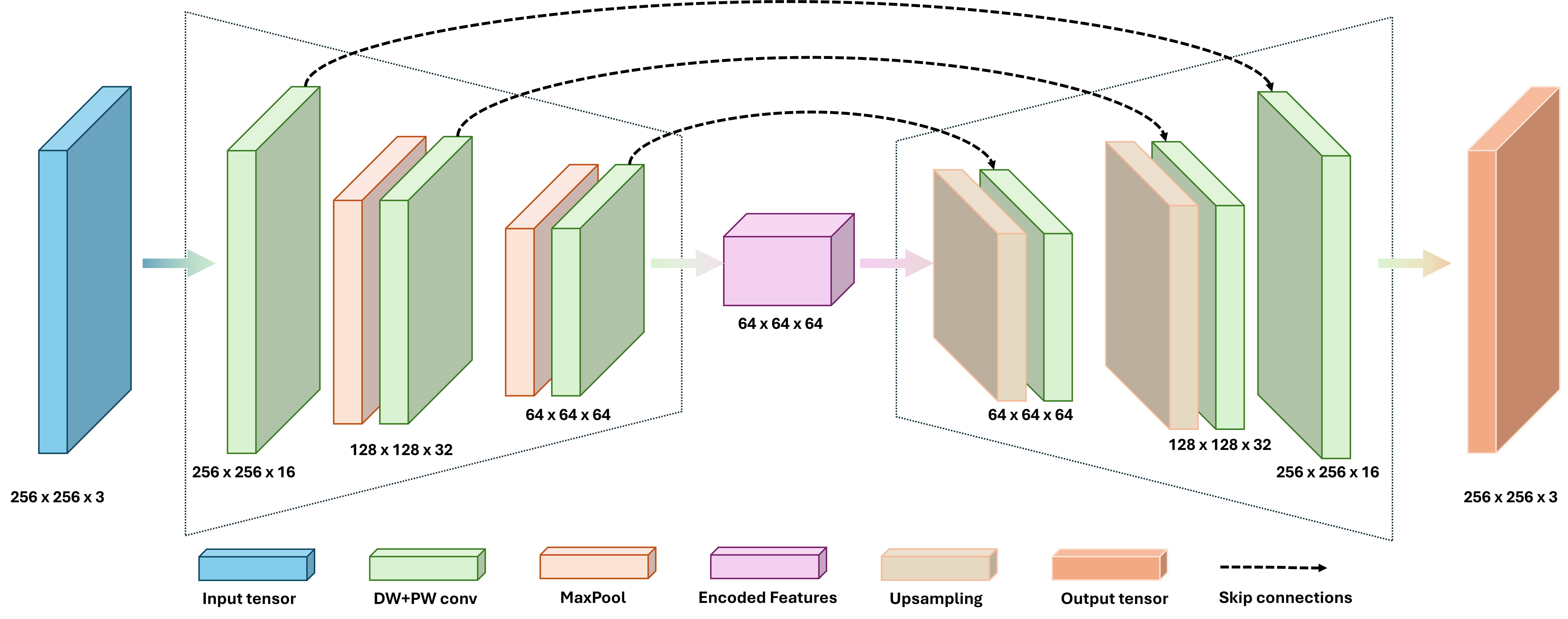}
    \caption{Proposed BioLite U-Net architecture with depthwise separable convolutions.}
    \label{fig:architecture}
\end{figure*}

As shown in Figure~\ref{fig:architecture}, the proposed BioLite U-Net comprises three key modules: an encoder, a bottleneck, and a decoder. The encoder gradually downsamples the input using two convolutional blocks, each consisting of depthwise separable convolutions, followed by max-pooling operations. These layers progressively capture hierarchical features while maintaining efficiency suitable for edge deployment.

The central bottleneck aggregates compressed high-level features and passes them to the decoder, which symmetrically upsamples the representation through bilinear interpolation and convolutional refinement layers. To mitigate the loss of spatial information during downsampling, skip connections are employed between encoder and decoder blocks at corresponding resolutions. These connections enable the network to recover fine-grained details, which are essential for segmenting small and thin structures such as ink traces and nozzle tips.

A final $1 \times 1$ convolution maps the decoded feature map to three output channels, followed by a softmax layer that generates per-pixel class probabilities for \textit{bioink}, \textit{nozzle}, and \textit{background}. This architecture is designed for rapid inference on low-power devices without significant compromise in segmentation accuracy.

\subsection{Training and Optimization Strategy}

To ensure robust and real-time segmentation performance, we design an efficient training pipeline tailored to our lightweight BioLite U-Net architecture. This subsection outlines the key aspects of the optimization strategy, including the loss function, optimizer settings, evaluation protocol, and comparison baselines.

\subsubsection{Loss Function}

We adopt the categorical cross-entropy loss to supervise pixel-wise multi-class segmentation. The loss is computed over all spatial locations $(i,j)$ in an image of size $H \times W$ with $C$ semantic classes:

\[
\mathcal{L}_{CE} = -\sum_{i=1}^{H}\sum_{j=1}^{W} \sum_{c=1}^{C} y_{i,j,c} \log(\hat{y}_{i,j,c})
\]

Here, $y_{i,j,c}$ represents the one-hot encoded ground truth label and $\hat{y}_{i,j,c}$ is the predicted softmax probability for class $c$ at pixel location $(i,j)$.

\subsubsection{Optimization and Training Setup}

Training is performed in PyTorch using the Adam optimizer, initialized with a learning rate of $1\times10^{-3}$ and a weight decay of $1\times10^{-5}$ to prevent overfitting. A mini-batch size of 4 was used throughout, and training was conducted for a total of 200 epochs to ensure convergence. The learning rate was scheduled to decay when validation performance plateaued, stabilizing the optimization process. 

\paragraph{Training Algorithm} The training pipeline is summarized in Algorithm~\ref{alg:training}.

\begin{algorithm}[ht]
\caption{Training Procedure for BioLite U-Net}
\label{alg:training}
\begin{algorithmic}[1]
\State \textbf{Input:} Training set $\mathcal{D}_{\text{train}}$, validation set $\mathcal{D}_{\text{val}}$, model $f_\theta$, learning rate $\eta$, batch size $B$
\State Initialize model parameters $\theta$
\State Initialize best validation loss: $\mathcal{L}_{\text{best}} \gets \infty$
\For{epoch = 1 to max\_epochs}
    \For{each mini-batch $(x, y)$ in $\mathcal{D}_{\text{train}}$}
        \State $\hat{y} \gets f_\theta(x)$ \hfill \textit{Forward pass}
        \State Compute loss $\mathcal{L}_{CE}(y, \hat{y})$
        \State Update $\theta \gets \theta - \eta \cdot \nabla_\theta \mathcal{L}_{CE}$
    \EndFor
    \State Evaluate validation loss $\mathcal{L}_{\text{val}}$ on $\mathcal{D}_{\text{val}}$
    \If{$\mathcal{L}_{\text{val}} < \mathcal{L}_{\text{best}}$}
        \State $\mathcal{L}_{\text{best}} \gets \mathcal{L}_{\text{val}}$
        \State Save current model parameters $\theta$
    \EndIf
\EndFor
\end{algorithmic}
\end{algorithm}

\subsubsection{Baseline Models}

To evaluate the performance of BioLite U-Net, we train and compare two additional segmentation baselines with lightweight encoders:

\begin{itemize}
    \item \textbf{MobileNetV2-DeepLab-like}: Uses MobileNetV2 as encoder with a reduced atrous spatial pyramid pooling (ASPP) decoder.
    \item \textbf{MobileNetV3-small}: Incorporates MobileNetV3-small as encoder with a lightweight decoder comprising bilinear upsampling and skip connections.
\end{itemize}

All models were trained and validated on the same dataset split, with 80\% of the images used for training, 10\% reserved for validation, and the remaining 10\% held out for testing. Performance was assessed using widely adopted segmentation metrics, namely pixel-wise accuracy, mean Intersection-over-Union (mIoU), and the Dice coefficient computed on a per-class basis to capture both global and class-specific segmentation quality.

\subsection{Deployment and Real-Time Inference}
\label{sec:deployment}

To enable practical real-time monitoring in bioprinting environments, we deploy the trained BioLite U-Net model directly on a Raspberry Pi 4 Model B, which features a 1.5 GHz quad-core CPU and 4 GB RAM. Despite its limited compute capacity, the lightweight design of our model ensures responsive inference performance. The high-level, real-time inference pipeline is described in Algorithm~\ref{alg:inference}.

The deployment process involves the following stages:

\begin{itemize}
\item \textbf{Model Export}: The PyTorch-trained model is exported using TorchScript tracing for compatibility with PyTorch Mobile runtime on ARM-based devices.
\item \textbf{Edge Runtime Setup}: The Raspberry Pi runs a minimal Python environment with OpenCV and Torch installed, enabling real-time frame capture, preprocessing, and inference.

\item \textbf{Inference Workflow}: Each captured frame from the bioprinting setup is resized to $256\times256$, normalized, and passed through the model to produce a segmentation mask.

\item \textbf{Latency Evaluation}: Inference latency is measured across 100 frames using CPU-only execution. The results demonstrate that the model maintains an average frame rate suitable for real-time monitoring applications without requiring GPU acceleration or quantization.
\end{itemize}

\begin{algorithm}[H]
\caption{Real-Time Inference on Raspberry Pi with BioLite U-Net}
\label{alg:inference}
\begin{algorithmic}[1]
\Require Input image $I \in \mathbb{R}^{H \times W \times 3}$, Traced model $\mathcal{M}{\text{Pi}}$
\Ensure Segmentation mask $M \in \mathbb{R}^{H \times W}$
\State $I{\text{resized}} \gets \text{Resize}(I, 256 \times 256)$
\State $I_{\text{norm}} \gets \text{Normalize}(I_{\text{resized}})$
\State $I_{\text{tensor}} \gets \text{ToTensor}(I_{\text{norm}})$
\State $I_{\text{input}} \gets \text{Unsqueeze}(I_{\text{tensor}})$ \Comment{Add batch dimension}
\State $\hat{Y} \gets \mathcal{M}{\text{Pi}}(I{\text{input}})$
\State $P \gets \text{Softmax}(\hat{Y})$
\State $M \gets \text{Argmax}(P, \text{axis}=1)$
\State \Return $M$
\end{algorithmic}
\end{algorithm}

\subsection{Evaluation Metrics}
\label{sec:evaluation}

To comprehensively evaluate the segmentation accuracy and deployment feasibility of the proposed framework, we used standard quantitative metrics.

\begin{itemize}
\item \textbf{Mean Intersection over Union (mIoU)}: Measures the average region overlap between predicted and ground truth masks across all classes. It is defined as:
\begin{equation}
\text{mIoU} = \frac{1}{C} \sum_{c=1}^{C} \frac{\mathrm{TP}_c}{\mathrm{TP}_c + \mathrm{FP}_c + \mathrm{FN}_c}
\end{equation}
where $\mathrm{TP}_c$, $\mathrm{FP}_c$, and $\mathrm{FN}_c$ denote the number of true positives, false positives, and false negatives for class $c$, respectively.
\item \textbf{Dice Score (F1)}: Captures the spatial overlap between the predicted segmentation and the ground truth, and is especially useful for imbalanced class distributions. For each class $c$:
\begin{equation}
    \text{Dice}_c = \frac{2 \cdot |\hat{Y}_c \cap Y_c|}{|\hat{Y}_c| + |Y_c|}
\end{equation}
where $\hat{Y}_c$ and $Y_c$ denote the predicted and ground truth binary masks for class $c$.

\item \textbf{Pixel Accuracy}: The overall proportion of correctly predicted pixels, calculated as:
\begin{equation}
    \text{Pixel Accuracy} = \frac{\sum_{i} \mathbb{1}[\hat{y}_i = y_i]}{N}
\end{equation}
where $\mathbb{1}$ is the indicator function, and $N$ is the total number of pixels.

\item \textbf{Inference Time (ms/frame)}: The average time required to process a single frame on the Raspberry Pi 4 Model B. This metric directly reflects the real-time capability of the deployed system.
\end{itemize}

All metrics are computed on the test set using the same preprocessing and image resolution. As discussed in Section~\ref{sec:results}, our BioLite U-Net achieves a strong balance between segmentation performance and low-latency inference suitable for embedded deployment in bioprinting.

\subsection{Model Complexity Comparison}
\label{sec:complexity}

To validate the suitability of our proposed BioLite U-Net for resource-constrained platforms, we compare its model complexity against two popular MobileNet-based segmentation baselines. Table~\ref{tab:complexity} summarizes the number of parameters, floating-point operations (FLOPs), and input image resolution used for each model.

\begin{table}[H]
\centering
\caption{Model complexity in terms of parameter count, FLOPs, and input resolution}
\label{tab:complexity}
\begin{tabular}{@{}lccc@{}}
\toprule
\textbf{Model} & \textbf{Parameters (M)} & \textbf{FLOPs (G)} & \textbf{Input Size} \\
\midrule
BioLite U-Net & 0.01 & 0.44 & 256×256 \\
MNetV2-DeepLabV3+ & 13.35 & 4.72 & 256×256 \\
MNetV3Small-FPN & 1.53 & 2.98 & 256×256 \\
\bottomrule
\end{tabular}
\end{table}

Despite using the same input resolution, BioLite U-Net achieves drastic reductions in model size and computational cost. Specifically, it contains only 0.01 million parameters and 0.44 GFLOPs per inference, which is over:

\begin{itemize}
\item \textbf{1300× smaller} in parameter count compared to MobileNetV2-DeepLabV3+.
\item \textbf{150× smaller} than MobileNetV3Small-FPN.
\item \textbf{10× - 11× fewer FLOPs} than either baseline.
\end{itemize}

These results confirm that BioLite U-Net is significantly more efficient, both in memory footprint and compute demand, making it highly suitable for real-time deployment on embedded devices like the Raspberry Pi 4B without requiring hardware accelerators or quantization.

\section{Dataset}
\label{sec:data}


To support the development and rigorous benchmarking of lightweight segmentation models for bioprinting, we created a dedicated, high-resolution, manually annotated dataset. The dataset comprises \textbf{787 RGB images} captured during real-time bioprinting operations.

\subsection{Capture Setup}
Images were collected using a Raspberry Pi 1.6-megapixel global shutter camera with a macro scale lens, positioned directly below the glass printing bed, focused on the tip of the extrusion needle. This configuration ensured consistently framed images while maintaining flexibility across varied printing conditions. The dataset was gathered during a wide range of bioprinting sessions using alginate-based bioinks, under ambient laboratory lighting. We deliberately varied extrusion speeds, nozzle geometries, and bioink types to capture rich, realistic variability. Each image prominently features the nozzle region, with fluctuating background elements and distinct bioink deposition patterns, ensuring robustness and real-world relevance.


\subsection{Class Definitions and Annotation}
\begin{figure}[htbp]
    \centering
    \includegraphics[width=0.9\linewidth]{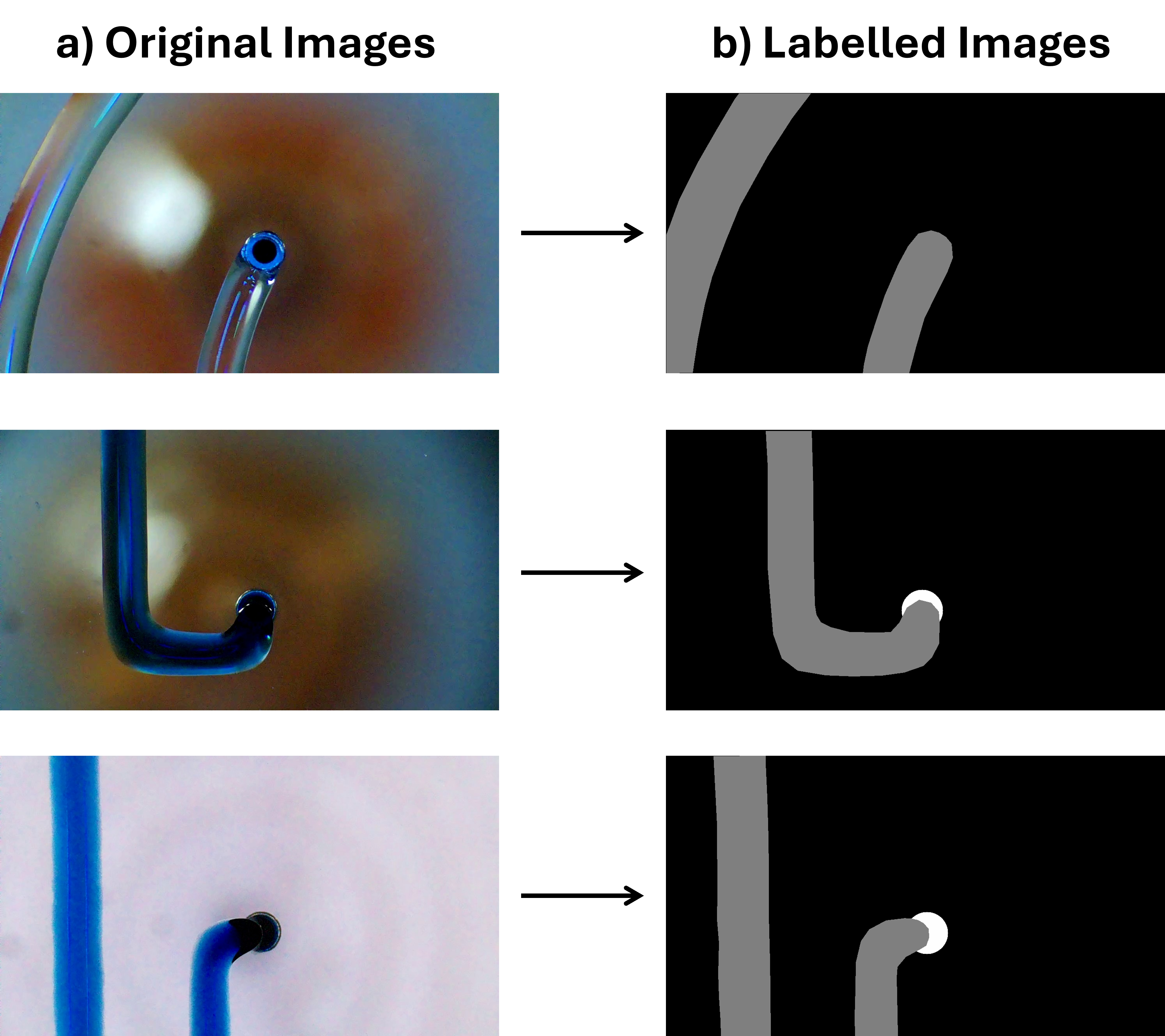}
    \caption{Example images from the bioprinting dataset.}
    \label{fig:dataset_example}
\end{figure}

Every image was meticulously labeled at the pixel level with one of three semantic classes:

\begin{itemize}
    \item \textbf{Background (Class 0)}: All other visual content, including the printer’s frame, print bed, and surrounding environment.
    \item \textbf{Bioink (Class 1)}: The extruded filament material.
    \item \textbf{Nozzle (Class 2)}: The heated metallic or polymer dispensing tip.
\end{itemize}
Annotations were manually generated using the VGG Annotation Software (VIA) \cite{via} and were independently cross-validated. An example from the dataset is shown in Figure~\ref{fig:dataset_example}.

\subsection{Data Split and Preprocessing}

The dataset was divided into fixed subsets to facilitate model development and unbiased assessment: 629 images (80\%) for training, 78 images (10\%) for validation, and 80 images (10\%) for testing. Each image was resized to a standardized 256×256 pixel resolution. During training, we applied a range of augmentation techniques to enhance generalization across lighting and mechanical variations:
\begin{itemize}
    \item Horizontal flips with 50\% probability
    \item Random brightness and contrast adjustment
    \item Random rotation between $-15^\circ$ and $15^\circ$
\end{itemize}

These augmentations significantly improve model robustness, especially when encountering subtle lighting shifts, motion artifacts, or diverse print configurations during deployment.

\subsection{Availability}

To promote openness and accelerate research in intelligent, vision-based bioprinting, we plan to release this dataset publicly upon publication. We believe it will serve as a valuable benchmark for lightweight, real-time semantic segmentation models and stimulate future work in biomedical fabrication, edge-AI, and automated tissue engineering.

\section{Experiments}

This section describes the experimental setup for training and evaluating the proposed BioLite U-Net architecture, as well as its comparison against MobileNet-based baselines. All experiments were conducted on our custom bioprinting dataset (Section \ref{sec:data}), with final evaluations performed on embedded hardware to assess deployment feasibility.

\subsection{Training Setup and Implementation Details}



All models were implemented in PyTorch 2.0 and trained on a workstation equipped with an NVIDIA RTX A6000 GPU (used exclusively for training). For deployment, the final PyTorch models were directly executed on a Raspberry Pi 4B with 4GB RAM, without any hardware accelerators or quantization, to ensure a realistic evaluation under constrained resources.

\begin{itemize}
    \item \textbf{Input size:} $256 \times 256$ RGB images
    \item \textbf{Loss function:} Categorical cross-entropy
    \item \textbf{Optimizer:} Adam ($\text{lr} = 10^{-3}$)
    \item \textbf{Batch size:} 4
    \item \textbf{Epochs:} 200 epochs
    \item \textbf{Regularization:} Weight decay of $10^{-5}$ and early stopping and checkpointing callbacks based on validation Dice score
\end{itemize}

We use random seed initialization and the same training/validation/test split across all models for fair comparison. All models were implemented in PyTorch 2.0. For MobileNet-based models, ImageNet-pretrained encoder weights were used, frozen for the first 20 epochs to stabilize convergence. BioLite U-Net was trained from scratch. No additional quantization or pruning was applied; all models were deployed in full-precision (FP32) on a Raspberry Pi 4B (4 GB RAM) without external accelerators. This setting reflects realistic embedded deployment constraints in robotic bioprinting systems.

\subsection{Models Evaluated}

We evaluated the following models:
\begin{itemize}
    \item \textbf{BioLite U-Net (ours)} – Custom encoder–decoder with depthwise separable convolutions and minimal parameter count.
    \item \textbf{MobileNetV2-DeepLabV3+} – MobileNetV2 encoder with a DeepLabV3+ decoder architecture.
    \item \textbf{MobileNetV3Small-FPN} – MobileNetV3-small backbone and simplified Feature Network Pyramid (FPN) decoder.
\end{itemize}

Our BioLite U-Net model was trained from scratch, while the MobileNet encoder models were loaded with pre-trained ImageNet weights.


    
    
    




\section{Results and Discussion}
\label{sec:results}

This section presents the quantitative and qualitative evaluation of the proposed BioLite U-Net architecture. We analyze segmentation performance on the test set and highlight the model’s runtime efficiency and compactness, comparing it with lightweight baselines such as MobileNetV2 and MobileNetV3.

\subsection{Training Results}

\begin{figure}[htbp]
\centering
\includegraphics[width=\linewidth]{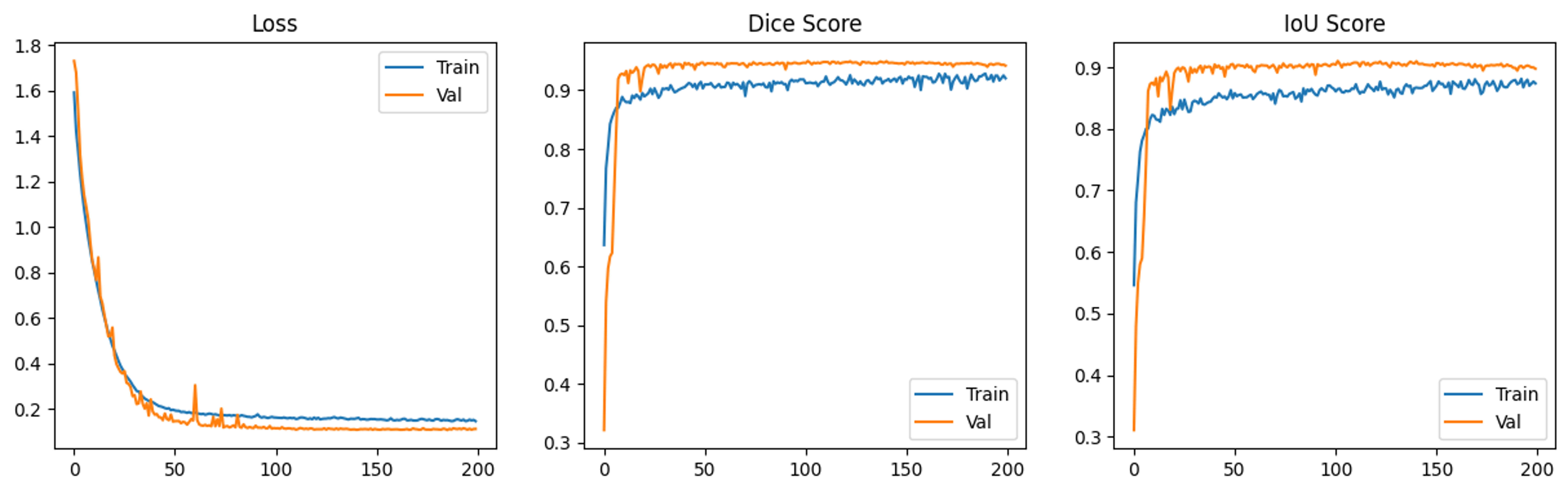}
\caption{Training curves of BioLite U-Net: loss, Dice, and IoU scores over 200 epochs.}
\label{fig:training_curves}
\end{figure}

Figure~\ref{fig:training_curves} illustrates the training and validation loss, Dice, and IoU over 200 epochs. The plots indicate strong generalization and stable convergence.

The validation performance quickly surpasses 90\% Dice within the first 20 epochs, with minimal overfitting. Both loss and metric curves remain consistent throughout training, suggesting effective optimization and model robustness.

\subsection{Quantitative Results}

Table~\ref{tab:results} reports the performance across key metrics: mean Intersection-over-Union (mIoU), Dice score, pixel accuracy, and inference time on a Raspberry Pi 4B device.

\begin{table}[H]
\centering
\caption{Comparison of segmentation performance on the test set using standard evaluation metrics}

\label{tab:results}
\begin{tabular}{lccc}
\toprule
\textbf{Model} & \textbf{mIoU (\%)} & \textbf{Dice (\%)} & \textbf{Pixel Acc. (\%)} \\
\midrule
BioLite U-Net (ours) & 92.85 & 96.17 & 99.55 \\
MNetV2-DeepLabV3+ & 94.33 & 97.03 & 99.55 \\
MNetV3Small-FPN & 91.68 & 95.52 & 99.42 \\

\bottomrule
\end{tabular}
\end{table}

The proposed BioLite U-Net achieves a Dice score of \textbf{96.17\%} and mIoU of \textbf{92.85\%}, outperforming both MobileNet baselines. Despite its compact design, it maintains segmentation quality while significantly reducing latency and compute requirements.

\begin{table}[H]
\centering
\caption{Inference Time (ms) on GPU and Raspberry Pi 4B}
\label{tab:inference_time_comparison}
\begin{tabular}{lcc}
\toprule
\textbf{Model} & \textbf{GPU (ms)} & \textbf{Raspberry Pi 4B (ms)} \\
\midrule
BioLite U-Net (ours) & 0.41 & 335 \\
MNetV2-DeepLabV3+ & 0.96 & 490.4 \\
MNetV3Small-FPN & 0.91 & 260.9 \\
\bottomrule
\end{tabular}
\end{table}

\textcolor{black}{
These results demonstrate that BioLite U-Net is well-suited for real-time edge deployment as well as for GPUs. Interestingly, lower FLOPs and parameter count do not always translate to lower CPU latency. On memory-limited devices such as the Raspberry Pi, skip connections can incur additional memory copies and cache misses, increasing runtime despite the model’s compactness. GPUs, by contrast, benefit from high bandwidth and parallelism, masking these penalties. }

\subsection{Qualitative Results}

\begin{figure}[htbp]
\centering
\includegraphics[width=\linewidth]{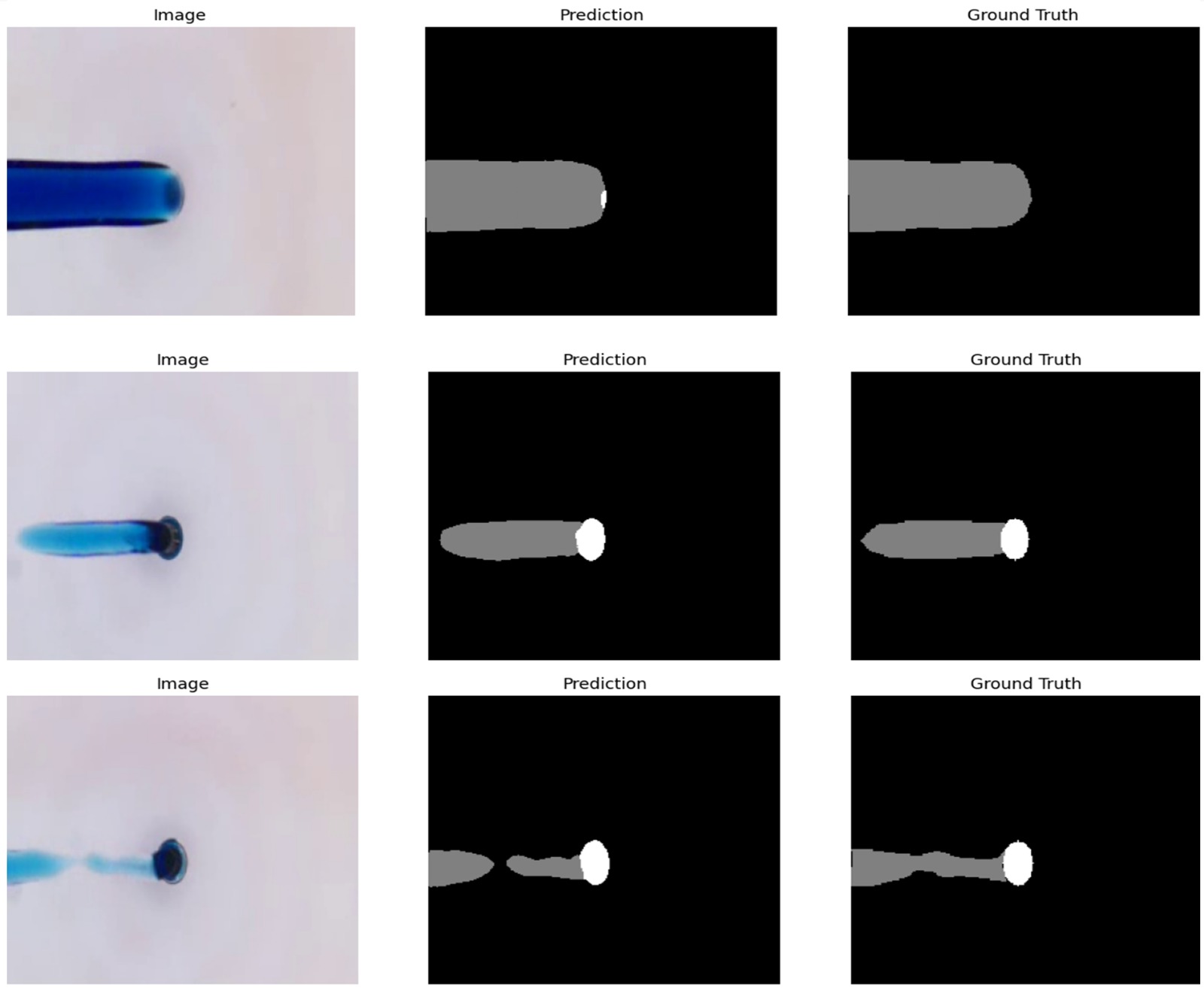}
\caption{Qualitative segmentation results of BioLite U-Net on test samples.}
\label{fig:qualitative}
\end{figure}

Figure~\ref{fig:qualitative} shows representative predictions on test images. The model successfully segments fine structures and preserves boundaries under varying illumination and texture conditions.

The predicted masks closely match ground truth labels, including in challenging areas such as object edges or filament transitions. However, minor mis-segmentation occurs in low-contrast regions.



    
    


\subsection{Discussion}

BioLite U-Net achieves a favorable balance between segmentation accuracy, inference speed, and model compactness, making it suitable for embedded robotics applications. Unlike MobileNet-based backbones, which were adapted from large-scale classification, BioLite U-Net is purpose-built for dense prediction in real time in the context of extrusion process dynamics. Depthwise separable convolutions reduce computational load, while skip connections ensure fine-grained spatial detail critical for nozzle and filament tracking. Another contributing factor is the encoder–decoder architecture, constructed with depthwise separable convolutions, which significantly reduces computation without sacrificing feature richness. Combined with skip connections, this structure ensures that low-level spatial details are preserved, which is critical for accurate segmentation of fine features such as nozzle tips and ink trails. 
From a deployment perspective, BioLite U-Net demonstrates competitive accuracy (within 2\% mIoU of MobileNetV2-DeepLabV3+) while requiring very few parameters and achieving $\sim$335 ms inference per frame on Raspberry Pi 4B. 
\textcolor{black}{While this is slower than the strict video frame rate definition of real-time, it is sufficient for bioprinting feedback, where extrusion dynamics occur over seconds rather than milliseconds. Comparable AI-based quality monitoring approaches in additive manufacturing report inference times in a similar range, e.g., 118.83 ms~\cite{zhu2024situ} and 461.8 ms~\cite{fu2023real}, supporting the practicality of our method for closed-loop bioprinting.} Furthermore, the model fits within the memory and compute constraints of commodity embedded boards, removing the need for external accelerators.
Key findings are summarized below:

\begin{itemize}
    \item \textbf{Real-time performance:} With an inference time of $\sim$335 ms, BioLite U-Net is well-suited for in-process print monitoring.
    
    \item \textbf{Competitive accuracy:} BioLite U-Net achieves only $\sim$2\% lower mIoU than the MobileNetV2-DeepLab baseline, while using over 1300$\times$ fewer parameters.

    \item \textbf{Robotics relevance:} Enables on-device decision-making for nozzle misalignment and extrusion faults, paving the way for closed-loop control.
    
    \item \textbf{Low resource footprint:} The model fits comfortably within the memory and compute constraints of embedded boards, even without hardware accelerators.

\end{itemize}

These trade-offs make BioLite U-Net an ideal choice for closed-loop control scenarios in bioprinting, where speed and on-device inference are crucial. For instance, the model can trigger corrective actions in real-time when the nozzle is misaligned or the extrusion flow is disrupted, thereby enhancing both print quality and operational robustness.

\subsection{Limitations and Future Work}

While our method demonstrates strong performance, several limitations remain that motivate future investigation:
\begin{itemize}
\item \textbf{Dataset size:} Limited to 787 annotated images under lab conditions. Broader generalization may require synthetic augmentation, domain randomization, or multi-view capture.
\item \textbf{Challenging substrates:} Performance may drop on reflective or transparent surfaces; future work could explore RGB-IR or depth sensor fusion.
\item \textbf{Temporal consistency:} Current inference is frame-based. Incorporating spatiotemporal models (e.g., ConvLSTM, 3D CNNs, attention mechanisms) could enhance robustness.
\item \textbf{Hardware optimization:} While real-time on Raspberry Pi 4B, throughput could benefit from hardware-aware NAS or compiler-level optimizations.
\end{itemize}
Beyond these, we plan to validate the model on additional bioink types, nozzle geometries, and multi-material bioprinting scenarios to assess cross-task generalization and robustness in diverse real-world conditions.

\section{Conclusion}

We introduced BioLite U-Net, a lightweight semantic segmentation framework for real-time monitoring in extrusion-based bioprinting. Alongside, we curated and annotated a dataset of 787 images, which, to our knowledge, is the first public dataset tailored to nozzle and bioink segmentation. The model is extremely compact (0.01M parameters, 0.44G FLOPs) yet achieves 92.85\% mIoU, 6.17\% Dice, and 99.55\% pixel accuracy on our curated dataset, closely matching MobileNet-based baselines despite being very small.

BioLite U-Net runs in 0.41 ms on a GPU and 335 ms on a Raspberry Pi 4B, demonstrating practical feasibility for embedded, low-power deployment. By combining efficiency and accuracy, it establishes the first segmentation framework tailored to bioprinting that is deployable on resource-constrained hardware, paving the way for closed-loop bioprinting systems with local vision-based feedback. Future work will extend to multi-material printing, sensor fusion, and temporal modeling.
\bibliographystyle{IEEEtran}
\bibliography{ref}

\end{document}